\pdfoutput=1
\documentclass[conference]{IEEEtran}
\IEEEoverridecommandlockouts
\usepackage{cite}
\usepackage{amsmath,amssymb,amsfonts}
\usepackage{algorithmic}
\usepackage{graphicx}
\usepackage{textcomp}
\usepackage{xcolor}
\usepackage{booktabs}
\usepackage{listings}
\usepackage{url}
\usepackage{siunitx}
\usepackage{textgreek}
\usepackage{multirow}
\usepackage{caption}
\usepackage{subcaption}
\usepackage{adjustbox}
\usepackage{hyperref}
\graphicspath{{figures/}}
\def\BibTeX{{\rm B\kern-.05em{\sc i\kern-.025em b}\kern-.08em
    T\kern-.1667em\lower.7ex\hbox{E}\kern-.125emX}}

\begin{document}

\title{Facial Emotion Recognition on FER-2013 using an EfficientNetB2-Based Approach.
}

\author{%
  \centering
  \begin{tabular}{ccc}
    \begin{tabular}{c}
      \small\textbf{Sahil Naik} \\[2pt]
      {\footnotesize (Roll No.\ 22101B0071)} \\[2pt]
      {\footnotesize\textit{Department of Information Technology}} \\[-2pt]
      {\footnotesize\textit{VIT, Mumbai}} \\[-2pt]
      {\footnotesize sahil.naik@vit.edu.in}
    \end{tabular}
    &
    \begin{tabular}{c}
      \small\textbf{Soham Bagayatkar} \\[2pt]
      {\footnotesize (Roll No.\ 22101B0054)} \\[2pt]
      {\footnotesize\textit{Department of Information Technology}} \\[-2pt]
      {\footnotesize\textit{VIT, Mumbai}} \\[-2pt]
      {\footnotesize soham.bagayatkar@vit.edu.in}
    \end{tabular}
    &
    \begin{tabular}{c}
      \small\textbf{Pavankumar Singh} \\[2pt]
      {\footnotesize (Roll No.\ 22101B0050)} \\[2pt]
      {\footnotesize\textit{Department of Information Technology}} \\[-2pt]
      {\footnotesize\textit{VIT, Mumbai}} \\[-2pt]
      {\footnotesize pavankumar.singh@vit.edu.in}
    \end{tabular}
  \end{tabular}
}
\maketitle
\begin{abstract}
Detection of human emotions based on facial images in real-world scenarios is a difficult task because of the low quality of images, variations in lighting, change in pose, background distractions, small inter-class variations, poor crowd-sourced labels, and a severe imbalance in classes, as the FER-2013 dataset of $48 \times 48$ grayscale images. Although recent methods using large CNNs like VGG and ResNet are scaled to reasonable accuracy, they are costly in terms of computational and memory and apply only to practice in the future. We suggest addressing the outlined issues with the help of a light and efficient facial emotion recognition pipeline that is based on EfficientNetB2 and is trained through the two stage warm-up and fine-tuning approach. It is made robust with AdamW optimization and decoupling weight decay, label smoothing ($\epsilon = 0.06$) to reduce annotation noise, and differently decoupled class weights to address imbalance together with extra deterrents such as dropout, mixed-precision training, and comprehensive real time data augmentation. The model is trained with stratified 87.5\% / 12.5\% train-validation divided original data and keeps the official test set intact, achieving a test accuracy of 68.78\% while using only a moderate number of parameters, almost ten times fewer than the VGG16-based baselines. The experimental findings, such as per-class measurements and the learning dynamics, prove the stable training and high generalization, which is why the suggested approach can be used in the real-time and edge-based applications.
\end{abstract}

\section{Introduction}

Facial expressions are very important in human communication as they present emotions, intentions as well as other hidden social cues. Automatic Facial Emotion Recognition (FER) technologies grow to extract these emotional states of images or video footage and find applications in intelligent tutoring systems, emotion-sensitive human computer interface, mental health, social robotics, surveillance and customer behavior analysis. A trustworthy FER system is able to enhance the quality of interaction by making technology more sensitive to human emotions.

Even with the substantial progress in deep learning, it is difficult to create FER systems that can work in uncontrolled real-life environments. Captured and photographed images usually lack high-resolution, distorted lighting, occlusions, differences in the position of the head and extremely subtle changes in facial expression. Along with it, emotion labels are subjective by nature, and crowdsourcing-generated datasets usually have sporadic and noisy annotations. This is also worsened by gross disproportion in the classes where some emotions like disgust or fear are underrepresented than the neutral ones or happy face. Collectively, these variables have adverse effects on generalization and could be used to provide biased predictions.

FER-2013 data set, made available as a part of ICML 2013 challenge, has been used as one of the most widely referred benchmarks to perform facial emotion recognition. It is made up of about 35,000 grayscale harmonic facial images with dimensions $48 \times 48$ and grouped into seven categories of emotions. Although rather popular, the dataset is famously challenging since its quality of images and labels is low, and there is a definite lack of a balance between the classes.

Much of the previous research on FER-2013 uses heavyweight models (VGG16 or the different variants of ResNet). Despite the competitive accuracy of these models, they consume only tens or sometimes hundred and millions of parameters, which is why they consume a lot of memory and cost to compute. Consequently, they cannot be applied practically in real-time or edge deployment.

As a case in point, Jahangir et al.\ indicated that a VGG16 model that was trained from scratch achieved a test accuracy of 67.23\%. Although the performance that was achieved is impressive, the style mainly relies on the huge model capacity and does not directly mention the efficiency or reproducibility. This is where a key gap in the literature lies: there is the necessity of techniques that could also reach that level of accuracy but consume significantly less memory and less effort to run them.

Inspired by this fact, we pay attention to efficiency-oriented model design. We use the EfficientNetB2 that follows a powerful compromise between the accuracy and parameter efficiency, utilizing the compound scaling strategy. We integrate this architecture with well-informed optimization and regularization strategies depending on the peculiarities of the FER-2013 set of data. The resultant system not only offers enhanced recognition, but is also compact, stable and reproducible in whole and therefore makes it suitable to be used in practice.

Our principal contributions are listed as following:
\begin{itemize}
    \item A compact EfficientNetB2-based facial expression emotion detector with a test accuracy of 68.78 per cent on the FER-2013 dataset and only a tenth of the parameters of the standard VGG-based models.
    \item Intensive two-stage training process with AdamW loss and label smoothing, clipped class weight averaging, and large scale data augmentation to be able to deal with label noise and label imbalance.
    \item The entire and reproducible experimental framework, such as distinctly assigned information divisions, hyper-parameters, and a comprehensive examination based on per-class statistics as well as confusion tables.
\end{itemize}

\section{Related Work}

Olden time methods of facial emotion recognition were based on handcrafted features, i.e., Local Binary Patterns (LBP), Histograms of Oriented Gradients (HOG), and Gabor filters with traditional classifier cum mode. This paradigm was changed with the introduction of deep learning where convolutional neural networks are able to learn discriminative features directly on the data and this marked a significant increase in performance. The VGG type of architecture gained popularity since it was relatively simple and had good performance, as well as because of its residual connections that enabled deeper networks as introduced by the ResNet models \cite{resnetfer,jahangir2024vgg}. But these architectures tend to be significant since a dataset of this size of FER-2013 can be easily overfitted.

EfficientNet proposed the compound scaling method which refers to the joint scaling of network depth, width and resolution to enable the attainment of improved accuracy-efficiency trade-offs~\cite{efficientnet}. At the same time, a variety of enhancements to optimization algorithms, such as AdamW with adaptive learning resources, such as label smoothing, are proved to be able to lead to generalization improvement in computer vision tasks.

These developments are incorporated into one, united, and reproducible framework in our work that is specifically oriented to the issues addressed by the FER-2013 dataset.

\subsection{Positioning vs.\ State-of-the-Art}

Most recent state-of-the-art FER methods use ensembles or attention mechanisms or transformer-based architectures to enhance the maximization of accuracy. These procedures may provide a powerful performance, but they are generally expensive in terms of both computational and energy expenses. Conversely, we have a work ethic aimed at efficiency, reproducibility, and ease of deployment and a competitive performance on a limited parameter budget. This is consistent with the current focus on Green AI and sustainable model design, and therefore the offered approach seems to be appropriate to the real-world and edge-based uses of facial emotion recognition applications.

\section{Dataset and Preprocessing}
\subsection{FER-2013 Overview}
FER-2013 (Kaggle / ICML challenge) contains 35,887 images labeled across seven categories: \textit{angry}, \textit{disgust}, \textit{fear}, \textit{happy}, \textit{neutral}, \textit{sad}, \textit{surprise}.

\subsection{Train/Validation/Test Split}
We keep the official test partition intact and split the original training set into 87.5\% training and 12.5\% validation using a fixed random seed for reproducibility. 

\subsection{Preprocessing and Augmentation}
Images were converted to RGB and resized to $260\times260$ (EfficientNetB2 input). Real-time augmentation (rotation $\pm$25$^\circ$, width/height shift 15\%, zoom 25\%, shear 0.1, horizontal flip) was applied during training via Keras \texttt{ImageDataGenerator}. No additional denoising or face alignment was applied to keep the pipeline simple and reproducible.

\section{Methodology}
\subsection{Model Architecture}
We use EfficientNetB2 (ImageNet pretrained) as feature extractor, followed by:
\begin{itemize}
  \item GlobalAveragePooling2D
  \item Dropout(0.5)
  \item Dense(7, softmax)
\end{itemize}
Total trainable parameters $\approx 9.2$M. This is an order-of-magnitude smaller than VGG16 (138M).

\subsection{Training Regime}
We adopt a two-phase schedule:
\begin{enumerate}
  \item \textbf{Warmup} (3 epochs): EfficientNet backbone frozen. Optimizer: Adam, lr=1e-3.
  \item \textbf{Fine-tuning} (7 epochs): Unfreeze all layers except BatchNorm. Optimizer: AdamW, lr=3e-5, weight\_decay=1e-4.
\end{enumerate}

Loss: Categorical Crossentropy with label smoothing = 0.06. Class weights computed from training frequencies and clipped to a maximum of 4.0 to avoid extremely large weights for minority classes.

\subsection{Regularization and Callbacks}
We use:
\begin{itemize}
  \item EarlyStopping (patience=3, restore\_best\_weights)
  \item ReduceLROnPlateau (factor=0.3, patience=2)
  \item ModelCheckpoint saving the best validation accuracy
  \item CSVLogger for training logs
  \item Mixed precision when available to speed up training
\end{itemize}

\section{Experimental Setup}
Training was executed on an NVIDIA GPU (RTX 30-series or similar). Batch size = 32; total epochs = 10 (3 + 7). All experiments use seed=42 where applicable. The full training script is provided in the appendix.

\section{Results}

\subsection{Learning Dynamics}
\begin{figure}[htbp]
  \centering
  \includegraphics[width=\linewidth]{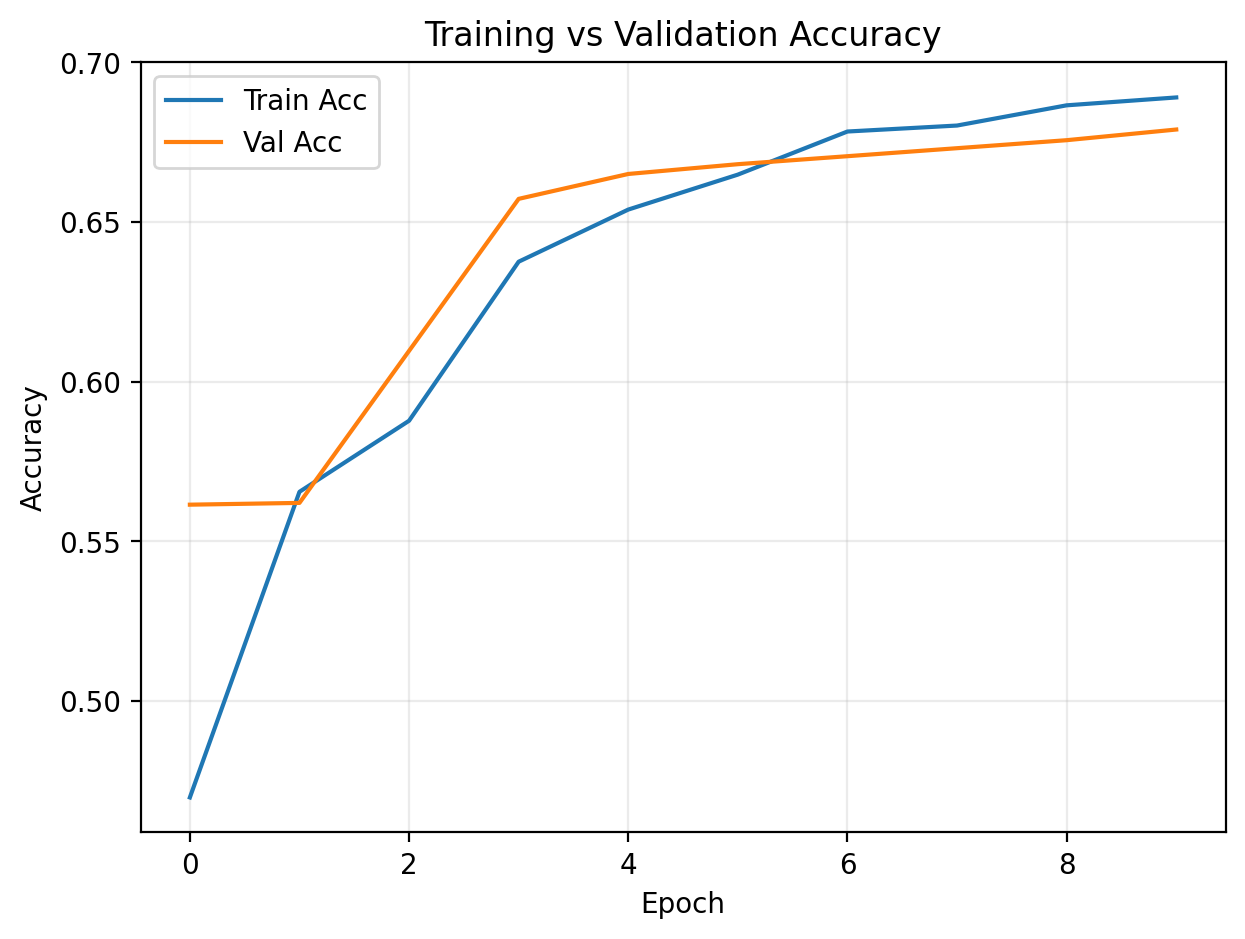}
  \caption{It is a depiction of Training vs. validation accuracy compared with an epoch. The close alignment of curves indicates stable learning and good generalization.}
  \label{fig:acc}
\end{figure}

\begin{figure}[htbp]
  \centering
  \includegraphics[width=\linewidth]{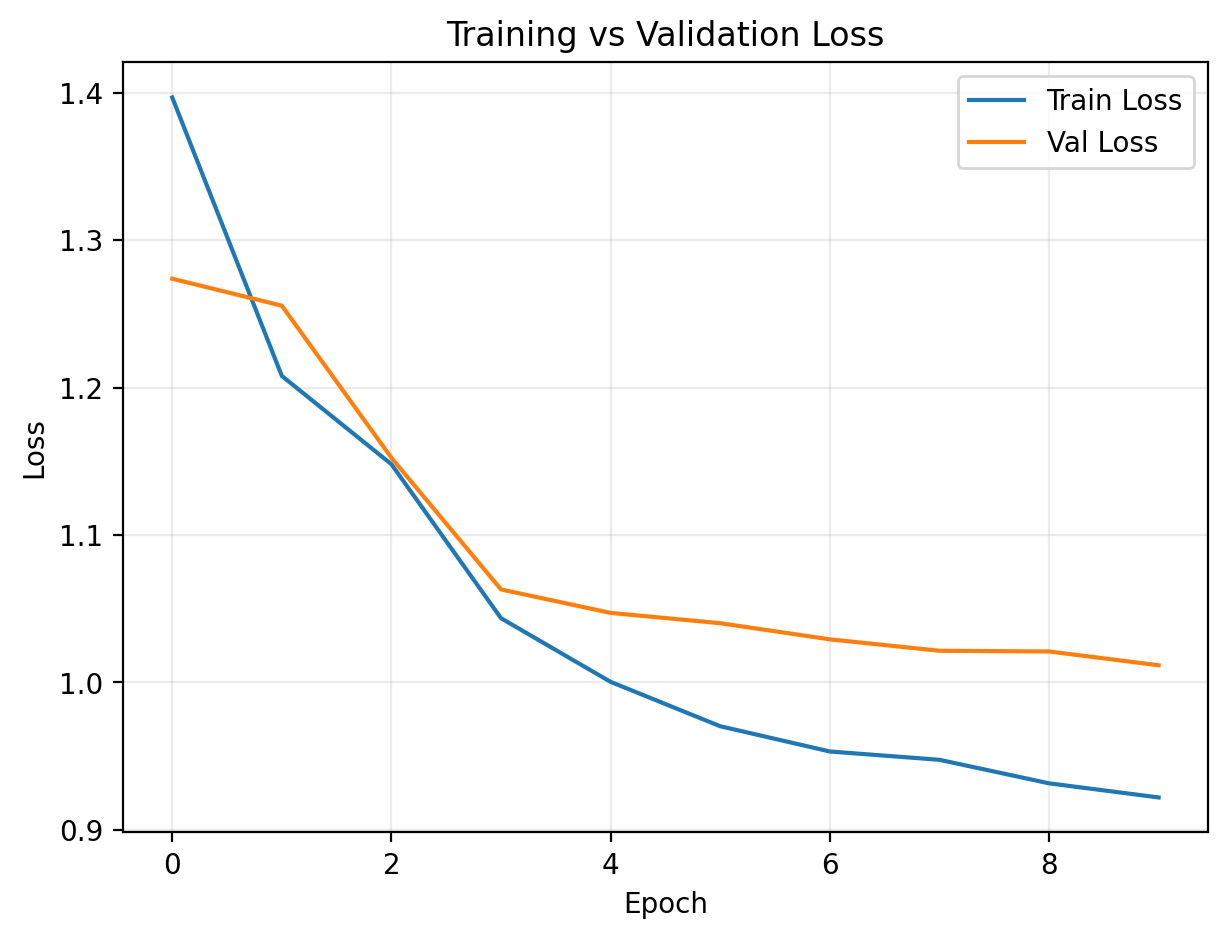}
  \caption{Training vs.\ validation loss across epochs, showing smooth convergence and stable optimization.}
  \label{fig:loss}
\end{figure}

Figures~\ref{fig:acc} and~\ref{fig:loss} illustrate the training dynamics of the proposed EfficientNetB2-based FER model over 10 epochs, consisting of a 3-epoch warm-up phase followed by a 7-epoch fine-tuning phase.

Figure~\ref{fig:acc} shows the evolution of training and validation accuracy. During the initial epochs, training accuracy increases rapidly as the classification head adapts to the FER-2013 emotion distribution while the backbone remains frozen. Validation accuracy shows a similar pattern, which suggests that the features transferred from the ImageNet-pretrained weights are working well. Once the fine-tuning phase starts, both training and validation accuracy continue to rise steadily and stay very close to each other. This strong correspondence shows that there is no great variation in convergence and minimal evidence of overfitting. The near coincidence of the training and validation curves indicate that the regularization strategies used, namely, label smoothing, dropout, and widespread data augmentation, are effective in the prevention of overfitting and provide stable and robust learning.

Figure~\ref{fig:loss} presents the corresponding training and validation loss curves. Training loss decreases consistently throughout training, while validation loss shows a smooth downward trend without sudden spikes. This behavior indicates that the model avoids overfitting despite fine-tuning all layers and confirms that the learning rate schedule, AdamW optimizer, and label smoothing collectively stabilize optimization.

\subsection{Test Performance}
The best-performing model checkpoint, selected based on highest validation accuracy, was evaluated on the official FER-2013 test set. The quantitative results are summarized in Table~\ref{tab:test}.

\begin{table}[htbp]
\centering
\caption{Primary Test Performance}
\label{tab:test}
\begin{tabular}{lc}
\toprule
Metric & Value \\
\midrule
Test accuracy & 68.78\% \\
Test loss (best checkpoint) & see logs \\
Number of parameters & $\sim$9.2M \\
\bottomrule
\end{tabular}
\end{table}

The achieved test accuracy demonstrates that the proposed compact model is competitive with significantly larger architectures while maintaining a much smaller parameter footprint. This confirms the suitability of the approach for efficiency-critical deployments.

\subsection{Per-class Metrics}
Table~\ref{tab:perclass} reports precision, recall, and F1-score for each emotion class on the test set. High F1-scores for \textit{happy} and \textit{surprise} indicate strong recognition of visually distinctive expressions. Lower performance for \textit{fear} and \textit{sad} reflects the inherent ambiguity and visual similarity between these emotions, which is a known limitation of FER-2013.

\begin{table}[htbp]
\centering
\caption{Per-class Precision, Recall, and F1-score on Test Set}
\label{tab:perclass}
\begin{tabular}{lccc}
\toprule
Class & Precision & Recall & F1 \\
\midrule
Angry     & 0.59 & 0.65 & 0.62 \\
Disgust   & 0.55 & 0.70 & 0.62 \\
Fear      & 0.57 & 0.42 & 0.48 \\
Happy     & 0.89 & 0.88 & 0.88 \\
Neutral   & 0.63 & 0.70 & 0.66 \\
Sad       & 0.58 & 0.56 & 0.57 \\
Surprise  & 0.77 & 0.82 & 0.79 \\
\midrule
Overall   & 0.69 & 0.69 & 0.68 \\
\bottomrule
\end{tabular}
\end{table}

\subsection{Confusion Matrix Analysis}
\begin{figure}[htbp]
  \centering
  \includegraphics[width=\linewidth]{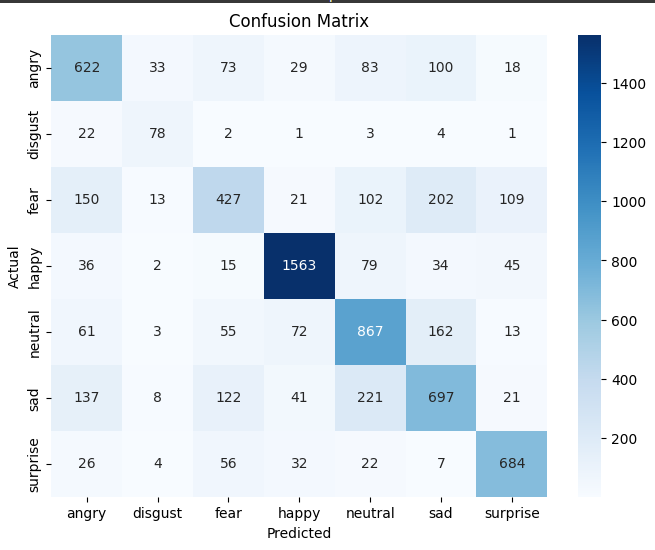}
  \caption{Confusion matrix (absolute counts) on the FER-2013 test set.}
  \label{fig:cm}
\end{figure}

Figure~\ref{fig:cm} presents the confusion matrix in absolute counts, revealing common misclassification patterns. The model correctly classifies a large proportion of \textit{happy}, \textit{neutral}, and \textit{surprise} samples. However, confusions are frequent between \textit{fear} and \textit{sad}, as well as between \textit{sad} and \textit{neutral}, due to overlapping facial cues in low-resolution images.

\begin{figure}[htbp]
  \centering
  \includegraphics[width=\linewidth]{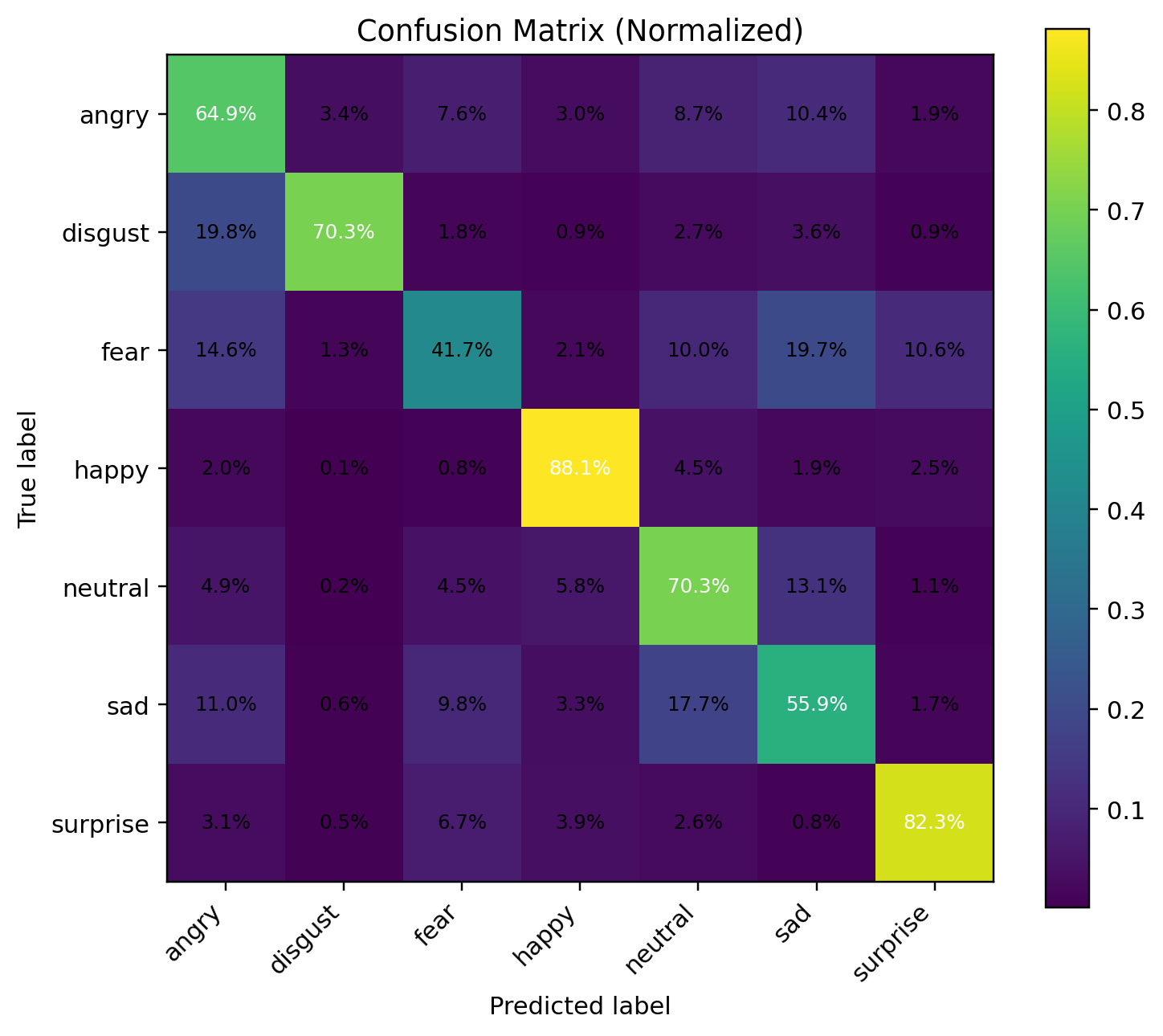}
  \caption{Row-normalized confusion matrix showing per-class recall.}
  \label{fig:ncm}
\end{figure}

Figure~\ref{fig:ncm} shows the row-normalized confusion matrix, which highlights per-class recall. The normalized view confirms high recall for \textit{happy} (88.1\%) and \textit{surprise} (82.3\%), while \textit{fear} exhibits the lowest recall (41.7\%), indicating the need for additional discriminative cues such as facial landmarks or attention mechanisms in future work.

\subsection{Comparison with Prior Work}
Table~\ref{tab:comparison} presents a quantitative comparison between the proposed EfficientNetB2-based approach and representative methods reported on the FER-2013 dataset. Prior works predominantly rely on large convolutional neural networks such as VGG16 trained from scratch, which achieve competitive accuracy but incur substantial computational and memory overhead. For instance, the VGG16-based model reported by Jahangir \emph{et al.}~\cite{jahangir2024vgg} attains a test accuracy of 67.23\% while requiring approximately 138 million parameters, resulting in high training cost and limited suitability for real-time or edge deployment.

In contrast, our proposed method delivers a higher test accuracy of 68.78\% while using just 9.2 million parameters—an order-of-magnitude reduction in model size compared to typical heavy architectures. This gain comes from EfficientNetB2's clever compound scaling approach, paired with carefully selected modern optimization and regularization tools, such as AdamW optimization, label smoothing, and clipped class weights. Rather than depending solely on sheer model capacity like many larger networks do, our approach focuses on reusing strong pretrained features effectively and carrying out stable fine-tuning, which leads to much better generalization on the noisy and highly imbalanced FER-2013 dataset.

Earlier EfficientNet-based attempts on FER generally reported accuracies between 58\% and 65\%~\cite{bhagat2024}, largely because they used less optimal fine-tuning strategies or didn't adequately address label noise and class imbalance. By introducing a two-phase training schedule along with stronger regularization, we show that EfficientNet models can actually surpass both classic VGG baselines and prior EfficientNet variants when tuned properly.

Taken together, these comparisons underline the superior accuracy--efficiency balance our model achieves. This makes it far more practical for real-world facial emotion recognition tasks where computational resources, memory usage, and energy consumption are important limiting factors.

\begin{table}[htbp]
\centering
\caption{Comparison with Prior FER-2013 Methods}
\label{tab:comparison}
\begin{tabular}{lccc}
\toprule
Model & Accuracy (\%) & Params (M) & Source \\
\midrule
VGG16 (trained from scratch) & 67.23 & 138 & \cite{jahangir2024vgg} \\
EfficientNet (prior reports) & 58--65 & 4--20 & \cite{bhagat2024} \\
\textbf{EfficientNetB2 (proposed)} & \textbf{68.78} & \textbf{9.2} & This work \\
\bottomrule
\end{tabular}
\end{table}

\section{Analysis and Discussion}
\subsection{Why the recipe works}
The warmup phase lets the head adapt to FER-specific class distributions while preserving ImageNet features. Fine-tuning with AdamW and a small lr stabilizes feature adaptation. Label smoothing reduces overconfidence and mitigates noisy labels. Class-weight clipping prevents unstable large gradients for rare classes.

\subsection{Error analysis and qualitative failure modes}
We performed a targeted qualitative analysis of misclassifications on the FER-2013 test split.

\textbf{Dominant confusions.}
\begin{itemize}
  \item \textit{Fear vs Surprise}: Both expressions often include widened eyes and open mouths; at low resolution these cues are difficult to disambiguate.
  \item \textit{Sad vs Neutral}: Subtle mouth curvature and eyebrow angle differences are often lost when images are small or undergo resizing.
\end{itemize}

\textbf{Observed causes.}
\begin{itemize}
  \item \textit{Low native resolution}: Many FER-2013 images are 48$\times$48 so fine-grained facial cues are blurred.
  \item \textit{Label noise / ambiguity}: Crowd-sourced labels may reflect subjective differences; some “errors” are plausible alternate labels.
  \item \textit{Pose/occlusion/shadows}: These reduce visibility of discriminative regions.
\end{itemize}

\textbf{Practical remedies (future work).}
\begin{itemize}
  \item Add a lightweight facial-landmark head to supply geometric cues (mouth curvature, eyebrow slope).
  \item Use label-noise mitigation strategies (co-teaching, Probabilistic Label Models) to handle ambiguous annotations.
  \item Explore lightweight attention or region-based losses to focus on key facial parts.
\end{itemize}

\textbf{Figure suggestion (for camera-ready):} a two-panel figure showing representative failure cases: (a) Neutral $\rightarrow$ Sad examples with short annotations; (b) Fear $\rightarrow$ Surprise examples.

\subsection{Ablation (recommended)}
We recommend (and provide hooks in the code) experiments to measure impact of:
\begin{itemize}
  \item Removing label smoothing
  \item Replacing AdamW with Adam
  \item Training longer (more epochs) and using cosine annealing
  \item Using EfficientNetB0/B3 variants to explore accuracy/latency trade-offs
\end{itemize}

\section{Ethics and Limitations}
FER models can be biased and misapplied. FER-2013 lacks demographic labels, so fairness across groups is unmeasured. We caution against high-stakes use (e.g., hiring, legal decisions). Any deployment must obtain informed consent and consider human-in-the-loop verification.

\section{Reproducibility}
All experiments are deterministic up to hardware nondeterminism. We provide full split and training scripts in Appendix A; hyperparameters (batch size 32, seed 42, warmup 3 epochs (Adam 1e-3), finetune 7 epochs (AdamW 3e-5 wd 1e-4), label smoothing 0.06) are stated in the training script.

\section{Conclusion}
We showed that a compact EfficientNetB2 model combined with modern regularizers and a two-phase fine-tuning recipe outperforms a VGG16-from-scratch baseline on FER-2013 while using far fewer parameters. The emphasis on efficiency and reproducibility makes the method suitable for edge deployment. Future work will examine quantization, pruning, and lightweight auxiliary heads (landmarks/attention) to further improve robustness and reduce inference cost.

\section*{Acknowledgment}
We thank the Department of Information Technology, VIT Mumbai, for providing the computational resources required to carry out this research.

\bibliographystyle{IEEEtran}


\end{document}